\documentclass[final,5p,times,twocolumn]{elsarticle}

\usepackage{amssymb}
\usepackage{amsmath}
\usepackage{booktabs}
\usepackage{multirow}
\usepackage{array}
\usepackage{tabularx}
\usepackage{graphicx}
\usepackage[table]{xcolor}
\usepackage{caption}
\usepackage[colorlinks=true,citecolor=cyan!55!black,linkcolor=cyan!55!black,urlcolor=cyan!55!black]{hyperref}
\hypersetup{
  pdfsubject={Preprint},
  pdfkeywords={Text-to-SQL, electronic medical records, chain of thought, schema graph, grounded evidence, large language models}
}

\newcolumntype{L}{>{\raggedright\arraybackslash}X}
\newcolumntype{P}[1]{>{\raggedright\arraybackslash}p{#1}}

\newcommand{\sql}[1]{\texttt{\spaceskip=0.5em plus 0.25em minus 0.05em\relax #1}}
\newcommand{\bad}[1]{\textcolor{red!70!black}{\textbf{#1}}}
\newcommand{\good}[1]{\textcolor{green!40!black}{\textbf{#1}}}
\newcommand{\evid}[1]{\textcolor{teal!70!black}{#1}}
\newcommand{\wempty}{w/\,$\emptyset$}
\newcommand{\woempty}{w/o\,$\emptyset$}
\newcommand{\cmark}{$\checkmark$}
\newcommand{\method}{VGSG-CoT}
\newsavebox{\caseboxL}
\newsavebox{\caseboxR}
\newlength{\caseheight}
\newcommand{\casepanel}[1]{\begin{minipage}[t]{0.475\textwidth}\raggedright #1\end{minipage}}
\newcommand{\caseframe}[1]{\fcolorbox{black!45}{black!3}{\parbox[t][\caseheight][t]{0.475\textwidth}{\usebox{#1}}}}
\newcommand{\casepair}[2]{%
  \sbox{\caseboxL}{\casepanel{#1}}\sbox{\caseboxR}{\casepanel{#2}}%
  \setlength{\caseheight}{\dimexpr\ht\caseboxL+\dp\caseboxL\relax}%
  \ifdim\dimexpr\ht\caseboxR+\dp\caseboxR\relax>\caseheight
    \setlength{\caseheight}{\dimexpr\ht\caseboxR+\dp\caseboxR\relax}\fi
  \caseframe{\caseboxL}\hfill\caseframe{\caseboxR}}

\journal{Knowledge-Based Systems}

\begin{document}

\begin{frontmatter}

\title{Value-grounded schema-graph reasoning for auditable Text-to-SQL in electronic medical records}

\author[inst1]{Weikuan Qiu\fnref{eq}}
\ead{202300400175@mail.sdu.edu.cn}
\author[inst1]{Chongzhao Xu\fnref{eq}}
\ead{xcz1404@qq.com}
\author[inst2]{Xijian Zhuo}
\ead{8206230419@csu.edu.cn}
\author[inst1]{Jinlian Ma\corref{cor1}}
\ead{jlma2019@sdu.edu.cn}
\ead{majinlian111@163.com}
\cortext[cor1]{Corresponding author.}
\fntext[eq]{These authors contributed equally to this work.}
\affiliation[inst1]{organization={School of Integrated Circuits, Shandong University},
            addressline={Shunhua Road Middle Section, High-tech Development Zone},
            city={Jinan},
            postcode={250101},
            country={China}}
\affiliation[inst2]{organization={School of Physics, Central South University},
            addressline={Yuelu District},
            city={Changsha},
            postcode={410000},
            country={China}}

\begin{abstract}
Electronic medical records (EMRs) store longitudinal patient data in relational databases, which are queried with structured query language (SQL) --- a language unfamiliar to most clinicians. While large language models (LLMs) can generate SQL from natural language questions, clinicians without SQL expertise still lack the means to audit how a generated query is grounded. We therefore address this gap with \method{} (Value-Grounded Schema-Graph Chain-of-Thought), a knowledge-grounded Text-to-SQL framework whose explicit chain of thought (CoT) exposes table, column, value, join, and aggregation decisions as auditable steps. Because explicit CoT can degrade accuracy in compact models, \method{} couples the CoT with a grounded evidence module (GEM) that retrieves value- and schema-level evidence from the database, and leverages grounded CoT distillation to train 1.5B and 7B generators that jointly produce the CoT and SQL. On MIMICSQL, \method{} achieves execution accuracies of 0.943 and 0.956 using the 1.5B and 7B generators, respectively, while the reasoning steps remain consistent with the final SQL. These results indicate that database-grounded evidence can offset the accuracy cost of explicit reasoning and enable auditable decision support for EMRs.
\end{abstract}

\begin{keyword}
Text-to-SQL \sep Electronic medical records \sep Chain of thought \sep Schema graph \sep Grounded evidence \sep Large language model
\end{keyword}

\end{frontmatter}

%% ======================================================================
\section{Introduction}
\label{sec:intro}

Electronic medical records (EMRs) constitute the primary evidence for clinical decision support, cohort discovery, and medical education \cite{johnson2016mimic,lee2022ehrsql}. Because their value grows with how easily they can be interrogated, timely access by clinicians directly affects the efficiency of diagnosis and treatment. In practice, EMR are stored in relational databases and retrieved with structured query language (SQL), which few medical professionals can write \cite{wang2020treqs,pan2021medts}. Clinical practice therefore needs systems that translate medical questions into correct SQL at an affordable computational cost. 

\begin{figure*}[t]
\centering
\includegraphics[width=\textwidth]{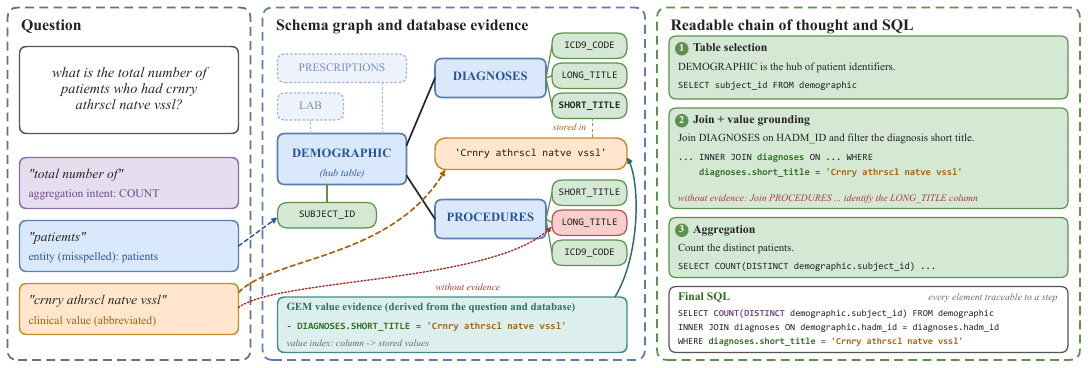}
\caption{An example of Text-to-SQL generation on EMR with an explicit chain of thought. Question spans are grounded to table, column, and value nodes of the schema graph, whose tables are linked by foreign keys on \sql{HADM\_ID}. Each reasoning step states one grounding decision in natural language together with an SQL fragment, so that the decision can be checked without writing SQL. Without database evidence, a 1.5B CoT model hosts the abbreviated diagnosis in \sql{PROCEDURES.LONG\_TITLE} (red); the value evidence of GEM identifies the column in which the value is stored.}
\label{fig:task}
\end{figure*}

Text-to-SQL maps a natural language question to an executable query over a database \cite{kim2020nl2sql,yu2018spider}. In Fig.~\ref{fig:task}, answering ``what is the total number of patients who had crnry athrscl natve vssl?'' requires grounding an abbreviated diagnosis to \sql{DIAGNOSES.SHORT\_TITLE}, joining the demographic hub table, and counting distinct patients. Unlike open-domain benchmarks \cite{yu2018spider,li2023bird}, medical questions are structurally shallow but semantically dense: in MIMICSQL, no query contains a nested sub-query or a set operation, whereas every query filters on a literal and many questions mention a value in a form that differs from the stored one.

However, Text-to-SQL systems face several shared problems \cite{kim2020nl2sql,liu2025survey}. The first is the ambiguity of natural language. Paraphrases, abbreviations, and misspellings relate only indirectly to schema elements, e.g., ``capso fungin'' denotes the drug caspofungin, and a mention may fit several tables, columns, or values, e.g., a diagnosis is recorded in both \sql{DEMOGRAPHIC.DIAGNOSIS} and \sql{DIAGNOSES.SHORT\_TITLE}, and a code of the International Classification of Diseases, Ninth Revision (ICD-9) may denote a diagnosis or a procedure. The second is schema complexity, i.e., recovering unstated join paths and storage conventions. The third concerns explainability and trustworthiness: a query is returned without an account of how it is constructed, so its user can only accept or reject it as a whole \cite{elgohary2020speak,tian2023steps}. Medicine amplifies the first and third problems, since one clinical concept has several coded and textual renderings, and an incorrect query still returns a plausible answer that may enter a clinical decision.

Large language models (LLMs) have changed the paradigm of Text-to-SQL: instruction following and a unified generation interface enable one model to read the schema, interpret the question, and emit SQL \cite{wei2022flan,hong2025survey,chen2026graph}. They also make it possible to expose the reasoning behind an answer. A chain of thought (CoT) \cite{wei2022cot} that presents intermediate decisions in a fixed structure has supported human interpretation and auditing in other domains \cite{alonso2026framing,zhou2026integrity}, and step-wise reasoning for natural-language-to-visualization generation lets users inspect the steps and locate errors \cite{shuai2025deepvis}. For EMRs, a CoT whose steps name the selected tables, columns, values, and joins gives medical professionals a readable interface to the query. Explicit reasoning, however, is not free. Its benefit grows with model scale \cite{wei2022cot}, and our analysis shows that CoT supervision alone lowers the execution accuracy of a compact model.

These grounding decisions depend on how the database stores medical concepts, which suggests drawing evidence from the database itself. Graph-empowered Text-to-SQL \cite{chen2026graph} encodes a schema graph into the attention of an LLM. SGU-SQL \cite{zhang2025sgusql} links question phrases to database elements through exact, partial, and value relations. SLM-SQL \cite{sheng2025slmsql} and FINER-SQL \cite{hoang2026finersql} demonstrate that compact models become competitive when trained with reasoning data or fine-grained execution feedback. Building on these observations, we render the value and column layers of the schema graph as textual evidence, so that each step of the CoT is grounded in stored values and schema elements that remain open to inspection.

We therefore propose \method{} (Value-Grounded Schema Graph with Chain-of-Thought), with three contributions: 
 \par (1) A structured CoT interface built on semantic-operator decomposition, which exposes each table, column, value, join, and aggregation decision as an auditable step and renders the reliability of these decisions quantifiable.
 \par (2) A grounded evidence module (GEM) that supplies value- and schema-level database evidence to offset the accuracy cost of explicit CoT, without reference queries, extra trainable parameters, or repeated LLM calls.
 \par (3) A grounded CoT distillation framework that trains compact generators to emit SQL with a consistent reasoning trace, combining deployment efficiency with auditability.
%With 1.5B and 7B backbones, \method{} reaches execution accuracies of 0.943 and 0.956 on MIMICSQL, respectively, while its CoT--SQL consistency exceeds 0.999.

%% ======================================================================
\section{Related work}
\label{sec:related}

\subsection{Question answering on electronic medical records}
 
Question answering (QA) over EMR covers unstructured and structured data. For clinical notes, emrQA \cite{pampari2018emrqa} formulates QA as machine reading comprehension, ignoring the relational organization of records. Structured EMRs QA is commonly cast as Text-to-SQL. Wang et al.~\cite{wang2020treqs} built MIMICSQL from the Medical Information Mart for Intensive Care III (MIMIC-III) \cite{johnson2016mimic} and proposed the Translate-Edit Model for Question-to-SQL (TREQS), a sequence-to-sequence model with condition-value recovery. MedTS \cite{pan2021medts} coupled BERT \cite{devlin2019bert} as the encoder with a grammar-based decoder; EHRSQL \cite{lee2022ehrsql} collected practical questions from hospital staff; and Tarbell et al.~\cite{tarbell2023medical} showed that models trained on template-derived questions generalize poorly beyond their templates. Graph-empowered Text-to-SQL \cite{chen2026graph} injected schema-graph information into an 8B LLM and reached 0.942 execution accuracy on MIMICSQL.

\subsection{Text-to-SQL generation}

Early natural language interfaces to databases relied on rules, templates, and ontologies \cite{androutsopoulos1995nlidb,saias2012binli}, which demand considerable manual effort for each domain \cite{kim2020nl2sql}. Neural models later learned the mapping end to end: SQLNet \cite{xu2017sqlnet} and RYANSQL \cite{choi2021ryansql} fill the slots of a SQL sketch; Coarse2Fine \cite{dong2018coarse} decodes a sketch before its details; and graph-based encoders, including Global-GNN \cite{bogin2019global}, graph neural schema encoding \cite{bogin2019gnn}, the relation-aware transformer RAT-SQL \cite{wang2020ratsql}, and Graphix-T5 \cite{li2023graphix}, model the schema as a graph to improve schema linking. Pre-trained language models \cite{devlin2019bert,liu2019roberta} further enabled parsers such as RESDSQL \cite{li2023resdsql}, which decouples schema linking from skeleton parsing.

LLMs introduced prompting and fine-tuning paradigms \cite{hong2025survey,liu2025survey}. Prompting methods select in-context examples \cite{gao2024dailsql,lee2025safe}, decompose questions \cite{pourreza2023dinsql,zhou2023least}, or align generation to reduce hallucination \cite{qu2024tasql}, while multi-agent systems such as contextual harnessing for efficient SQL synthesis (CHESS) \cite{talaei2024chess} and CHASE-SQL \cite{pourreza2025chase} add schema pruning and candidate selection. Fine-tuning methods adapt open models with synthesized data, e.g., OmniSQL \cite{li2025omnisql}, BASE-SQL \cite{sheng2025basesql}, and SPFT-SQL \cite{zhang2025spft}. LearNAT \cite{liao2026learnat} is closest to our data pipeline: it decomposes queries into executable sub-queries by Monte Carlo tree search (MCTS) \cite{browne2012mcts} guided by abstract syntax trees (ASTs) and aligns the model with direct preference optimization (DPO) \cite{rafailov2023dpo}. We retain AST verification, but replace the decomposition unit with semantic operators to suit medical cases. Database content is a further source of evidence: BRIDGE \cite{lin2020bridge} appends cell values that match the question; and CodeS \cite{li2024codes} and CHESS \cite{talaei2024chess} retrieve stored values for the prompt.

\subsection{Reliability of generated explanations}

CoT prompting elicits intermediate reasoning steps \cite{wei2022cot}; and decomposition methods such as least-to-most prompting \cite{zhou2023least} and DIN-SQL \cite{pourreza2023dinsql} solve Text-to-SQL by breaking the task into sub-problems. Explicit reasoning has also served as an interface for people. DeepVIS \cite{shuai2025deepvis} equips natural-language-to-visualization generation with step-wise reasoning and an interface in which users inspect the steps, identify errors, and adjust the output. For Text-to-SQL, interactive systems explain queries to non-expert users and let them correct the explanation in natural language \cite{elgohary2020speak} or edit a step-by-step explanation directly \cite{tian2023steps}. Structured CoT has supported human interpretation of subjective classifications \cite{alonso2026framing}; and structured prompts with evidence-grounding and traceability constraints have improved the integrity of CoT in security analysis \cite{zhou2026integrity}.

The reliability of an explanation comprises several properties that should not be conflated. Consistency asks whether the explanation entails the prediction, and factuality whether it is supported by the input evidence \cite{ye2022unreliability}. Faithfulness, i.e., whether the explanation reflects the computation that produced the answer, is stricter \cite{jacovi2020faithful}. CoT explanations can be plausible yet unfaithful \cite{turpin2023unfaithful}, which motivated faithful CoT \cite{lyu2023faithful} to execute a symbolic chain with a deterministic solver. In addition, the benefit of CoT depends on model scale \cite{wei2022cot}; training small models on rationales narrows this gap \cite{hsieh2023distilling,sheng2025slmsql}. However, as observed in our experiments, explicit steps still incur accuracy costs at the 1.5B scale. Our CoT pairs each step with an executable SQL fragment, and we measure its consistency with the final SQL and its factuality with respect to the database.

%% ======================================================================
\section{Datasets}
\label{sec:data}

We evaluated on MIMICSQL, the target domain of this work, and on Spider for cross-domain generalization. Dataset statistics are computed over their combined training, development, and test splits.

\begin{table}[t]
\caption{The summary of MIMICSQL dataset.}
\label{tab:mimicsql}
\footnotesize
\begin{tabular*}{\columnwidth}{@{\extracolsep{\fill}}lr@{}}
\toprule
Data & Value\\
\midrule
Number of patients & 46\,520\\
Number of tables & 5\\
Number of columns in tables & 23/5/5/8/9$^{\mathrm{a}}$\\
Number of question--SQL pairs & 10\,000\\
Training/validation/test pairs & 8\,000/1\,000/1\,000\\
Average template question length (in words) & 18.39\\
Average natural language question length (in words) & 16.45\\
Average SQL query length & 21.14\\
Average aggregation columns & 1.10\\
Average conditions & 1.76\\
\bottomrule
\end{tabular*}

\vspace{2pt}
{\scriptsize $^{\mathrm{a}}$ Demographics, Diagnoses, Procedures, Prescriptions, and Laboratory tests, counted on the released database schema.\par}
\end{table}

\textbf{MIMICSQL.} MIMICSQL \cite{wang2020treqs} is the first large-scale Text-to-SQL dataset on EMR. It is derived from MIMIC-III \cite{johnson2016mimic}, a de-identified critical care database, from which Wang et al.~\cite{wang2020treqs} extracted five tables: demographics (\sql{DEMOGRAPHIC}), diagnoses (\sql{DIAGNOSES}), procedures (\sql{PROCEDURES}), prescriptions (\sql{PRESCRIPTIONS}), and laboratory tests (\sql{LAB}). The tables form a hub-and-spoke schema in which \sql{DEMOGRAPHIC} stores one row per admission and each clinical table refers to it through the admission identifier \sql{HADM\_ID} and the patient identifier \sql{SUBJECT\_ID}. Table~\ref{tab:mimicsql} summarizes the dataset.

The question--SQL pairs are collected in two stages \cite{wang2020treqs}. First, template questions are generated for two intents: retrieval questions ask for attributes of a patient, disease, procedure, prescription, or laboratory test; and reasoning questions ask for the number of patients or for the maximum, minimum, or average of an attribute under given conditions. Each template instantiates the query form \sql{SELECT \$AGG\_OP(\$AGG\_COLUMN) FROM \$TABLE WHERE \$COND\_COLUMN \$COND\_OP \$COND\_VAL}, where the aggregation operator is \sql{COUNT}, \sql{MAX}, \sql{MIN}, \sql{AVG}, or none and the condition operator is one of $=$, $<$, $\leq$, $>$, and $\geq$. Second, eight freelancers with medical knowledge remove unrealistic template questions and paraphrased the others into natural language. Following previous work \cite{wang2020treqs,pan2021medts,chen2026graph}, we used the natural language questions and the official random split into 8,000 training, 1,000 validation, and 1,000 test pairs.

\textbf{Spider.} Spider \cite{yu2018spider} is a cross-domain benchmark whose questions and queries are annotated over databases from many domains. Its released training, development, and test sets contain 11,840 questions over 206 databases, and the databases of the three sets are disjoint. We used Spider to examine whether the CoT interface and GEM transfer to databases whose schemas, values, and query structures differ from those of EMR.

\textbf{Query hardness.} Following Spider \cite{yu2018spider}, we assigned each query to one of four hardness levels using three counts: $A$, the number of keywords from $K_1=\{$\sql{WHERE}, \sql{GROUP BY}, \sql{ORDER BY}, \sql{LIMIT}, \sql{JOIN}, \sql{OR}, \sql{LIKE}, \sql{HAVING}$\}$; $B$, the number of keywords from $K_2=\{$\sql{EXCEPT}, \sql{UNION}, \sql{INTERSECT}, \sql{NESTED}$\}$; and $C$, the number of satisfied rules in $R_1$, namely more than one aggregation function, selected column, condition, or \sql{GROUP BY} clause. A query is easy if $A\leq1$ and $B=C=0$; medium if $B=0$ and either $A\leq1, C\leq2$ or $A\leq2, C<2$; hard if $B=0$ and either $A\leq2, C>2$ or $2<A\leq3, C\leq2$, or if $A\leq1$, $B=1$, and $C=0$; and extra hard otherwise. Table~\ref{tab:stats} reports the distribution on both datasets.

\begin{table}[t]
\caption{Structural and semantic statistics of MIMICSQL and Spider, computed on the union of the training, development, and test sets of each dataset.}
\label{tab:stats}
\footnotesize
\begin{tabular*}{\columnwidth}{@{\extracolsep{\fill}}lrr@{}}
\toprule
Statistic & MIMICSQL & Spider\\
\midrule
Questions / databases & 10\,000 / 1 & 11\,840 / 206\\
Hardness: easy & 742 & 2\,701\\
\quad medium & 8\,236 & 4\,302\\
\quad hard & 1\,022 & 2\,559\\
\quad extra hard & 0 & 2\,278\\
Tables per query: one & 3\,573 & 6\,254\\
\quad two & 5\,405 & 3\,815\\
\quad three or more & 1\,022 & 1\,771\\
Aggregation: \sql{COUNT} & 7\,687 & 2\,502\\
\quad other aggregation only & 715 & 1\,024\\
\quad none & 1\,598 & 8\,314\\
Nested sub-query or set operation & 0 & 1\,855\\
At least one literal condition & 10\,000 & 6\,385\\
Clinical-vocabulary columns constrained$^{\mathrm{a}}$ & 6\,697 & --\\
Stored value not mentioned verbatim$^{\mathrm{b}}$ & 2\,251 & 696\\
\bottomrule
\end{tabular*}

\vspace{2pt}
{\scriptsize $^{\mathrm{a}}$ \sql{SHORT\_TITLE}, \sql{LONG\_TITLE}, \sql{ICD9\_CODE}, \sql{DRUG}, \sql{DRUG\_TYPE}, \sql{FORMULARY\_DRUG\_CD}, \sql{ROUTE}, \sql{LABEL}, \sql{FLUID}, \sql{CATEGORY}, and \sql{DEMOGRAPHIC.DIAGNOSIS}. $^{\mathrm{b}}$ Case-insensitive substring test between each literal of the reference query and the question.\par}
\end{table}

\textbf{Semantic density.} Table~\ref{tab:stats} shows that MIMICSQL is structurally shallow: none of its queries contains a nested sub-query or a set operation, none is extra hard, and 89.8\% involve at most two tables. In contrast, 15.7\% of the Spider queries are nested or combine sub-queries by set operations and 19.2\% are extra hard. The difficulty of MIMICSQL is semantic instead. Every query filters on at least one literal, against 53.9\% in Spider; 67.0\% of the queries constrain clinical-vocabulary columns such as \sql{SHORT\_TITLE}, \sql{LONG\_TITLE}, \sql{ICD9\_CODE}, \sql{DRUG}, or \sql{LABEL}; and 22.5\% of the questions mention at least one value in a form that differs from the stored one, against 5.9\% in Spider, owing to abbreviations, reordering, misspellings, number formats, or implicit codes such as ``dead'' for \sql{EXPIRE\_FLAG = 1}. Moreover, one concept is often recorded in several columns, e.g., a diagnosis in \sql{DEMOGRAPHIC.DIAGNOSIS}, \sql{DIAGNOSES.SHORT\_TITLE}, and \sql{DIAGNOSES.LONG\_TITLE}. Owing to their shallow syntactic structure, medical-domain questions are prone to having their true grounding difficulty underestimated. This observation motivates both our semantic-operator CoT and the value- and schema-level evidence provided by GEM.

%% ======================================================================
\section{Method}
\label{sec:method}

\begin{figure*}[t]
\centering
\includegraphics[width=\textwidth]{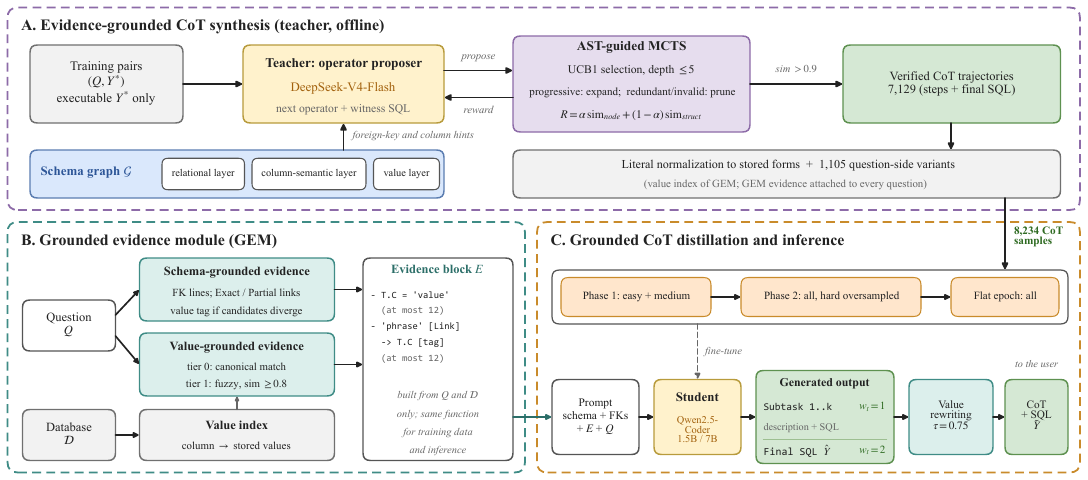}
\caption{Overview of \method. (A) Evidence-grounded CoT synthesis: a teacher LLM proposes semantic operators that AST-guided MCTS verifies against the reference query. (B) GEM derives value- and schema-grounded evidence from the question and the database with the same function for training data and inference. (C) Grounded CoT distillation trains a compact student on the CoT samples; at inference, the student emits the CoT and the final SQL, and value rewriting restores the stored forms of literals. FK: foreign key.}
\label{fig:overview}
\end{figure*}

Fig.~\ref{fig:overview} shows the overall architecture of \method. Its output is a structured CoT together with the final SQL. The GEM derives value- and schema-level evidence from the database, evidence-grounded synthesis produces verified CoT trajectories with a teacher LLM, and grounded CoT distillation trains a compact student that generates the CoT and the SQL with this evidence in its prompt.

\subsection{Problem formulation}
\label{sec:problem}

Let $\mathcal{D}$ be a relational database with schema $\mathcal{S}=(\mathcal{T},\mathcal{C},\mathcal{F})$, where $\mathcal{T}$, $\mathcal{C}$, and $\mathcal{F}$ denote its tables, columns, and foreign keys, respectively. Let $\mathcal{V}(c)$ denote the values stored in column $c\in\mathcal{C}$. A conventional Text-to-SQL system maps a question $Q$ and $\mathcal{S}$ to a query $Y$. We instead require two outputs,
\begin{equation}
(Q,\mathcal{S},\mathcal{D})\rightarrow(Z,Y),
\label{eq:task}
\end{equation}
where $Z=(z_1,\dots,z_k)$ is a structured CoT and $Y$ is the final SQL. Two objectives follow. The first is execution correctness, $[\![Y]\!]_{\mathcal{D}}=[\![Y^{*}]\!]_{\mathcal{D}}$, where $Y^{*}$ is the reference query. The second is CoT--SQL consistency, $\mathrm{Con}(Z,Y)=1$, i.e., every table, column, value, aggregation, and join of $Y$ is declared in some step of $Z$, and every element declared in $Z$ occurs in $Y$.

Query construction can be decomposed into grounding decisions over tables, columns, literals, aggregation operators, and join paths. The schema graph $\mathcal{G}$ represents the evidence required to make these decisions through three layers: a relational layer linking tables by foreign keys, a column-semantic layer linking medical concepts to columns, and a value layer linking stored values to their columns. GEM renders evidence $E$ from $\mathcal{G}$ and $\mathcal{D}$ during both training and inference. With this evidence as input, the generator follows the conditional distribution
\begin{equation}
p_{\theta}(Z,\hat{Y}\mid Q,\mathcal{S},E),
\label{eq:gen}
\end{equation}
where $\theta$ denotes all trainable parameters of the model. $\hat{Y}$ denotes the SQL statement generated by the model.
\subsection{Structured CoT as an auditable interface}
\label{sec:cot}

\begin{table*}[t]
\caption{Decomposition logic of the five semantic operators. Each step records one grounding decision together with an SQL fragment that witnesses it.}
\label{tab:operators}
\footnotesize
\begin{tabularx}{\textwidth}{@{}P{2.0cm}P{3.5cm}LP{3.3cm}@{}}
\toprule
Operator & Grounding decision (graph layer) & Witness SQL fragment & Ordering constraint\\
\midrule
Table selection & Entity $\rightarrow$ table (relational) & \sql{SELECT DEMOGRAPHIC.SUBJECT\_ID FROM DEMOGRAPHIC} & First step; hub table before clinical tables\\
Column identification & Concept $\rightarrow$ column (column-semantic) & \sql{SELECT DIAGNOSES.SHORT\_TITLE FROM DIAGNOSES} & After the host table is grounded\\
Value grounding & Surface form $\rightarrow$ stored value (value) & \sql{... WHERE DIAGNOSES.SHORT\_TITLE = 'Crnry athrscl natve vssl'} & After its column; hub conditions before joins\\
Join & Table pair $\rightarrow$ foreign-key path (relational) & \sql{... INNER JOIN DIAGNOSES ON DEMOGRAPHIC.HADM\_ID = DIAGNOSES.HADM\_ID} & One step per clinical table\\
Aggregation & Intent $\rightarrow$ \sql{COUNT}, \sql{MAX}, \sql{MIN}, \sql{AVG}, or none & \sql{SELECT COUNT(DISTINCT DEMOGRAPHIC.SUBJECT\_ID) ...} & Last step, after all grounding\\
\midrule
\multicolumn{4}{@{}l}{\textit{Composition patterns}: single table: table + column $\rightarrow$ aggregation; two tables: hub $\rightarrow$ join + column + value $\rightarrow$ aggregation;}\\
\multicolumn{4}{@{}l}{multi-hop: hub $\rightarrow$ join + value $\rightarrow$ join + value $\rightarrow$ aggregation.}\\
\bottomrule
\end{tabularx}
\end{table*}

LearNAT \cite{liao2026learnat} decomposes queries into executable sub-queries, suiting structurally complex benchmarks. Since MIMICSQL is shallow in structure but dense in semantics (Section~\ref{sec:data}), we took the semantic operator, i.e., one grounding decision, as the unit of a step. Table~\ref{tab:operators} lists the five operators and their ordering constraints. Each step $z_i=(d_i,s_i)$ pairs a natural language description $d_i$ with an SQL fragment $s_i$ that witnesses the decision (Fig.~\ref{fig:task}). The description lets a reader follow how the query is constructed without writing SQL, and the fragment ties the description to the executed query, which makes the consistency between $Z$ and $Y$ measurable.

\subsection{Grounded evidence module}
\label{sec:gem}

GEM renders the value and schema layers of $\mathcal{G}$ as an evidence block $E$ in the prompt (Fig.~\ref{fig:overview}B). It reads only the question and the database and has no trainable parameters. The same function renders $E$ for the training data and at inference, and a self-check confirms that re-rendered evidence is byte-identical to the stored training evidence.

\subsubsection{Value-grounded evidence}
\label{sec:vg}

GEM indexes $\mathcal{D}$ once by collecting the distinct stored values of every column, keyed by their canonical forms, i.e., upper-cased strings without spaces and punctuation. Contiguous word $n$-grams of $Q$ with $n\leq5$ are matched against the index, longer spans first. A span whose canonical form equals that of a stored value yields an exact hit; otherwise, values found by containment, an edit distance of one, or character-trigram overlap yield a fuzzy hit when their similarity to the span reaches 0.8. At most 12 hits are rendered as lines \sql{T.C = 'v'}, which state both the stored spelling of a value and its host column.

After generation, value rewriting $\tilde{Y}=r(\hat{Y},\mathcal{D})$ normalizes escaped apostrophes to the SQLite convention and examines each literal $\ell$ compared with a column $c$ in $\hat{Y}$. If $\ell\notin\mathcal{V}(c)$, it is replaced by
\begin{equation}
\begin{aligned}
v^{*}&=\mathop{\arg\max}_{v\in\mathcal{V}(c)}\ \mathrm{sim}(\ell,v),\\
\mathrm{sim}(\ell,v)&=\max\Big(\mathrm{Dice}_{3}(\ell,v),\,1-\frac{\mathrm{ED}(\ell,v)}{\max(|\ell|,|v|)}\Big),
\end{aligned}
\label{eq:rewrite}
\end{equation}
provided that $\mathrm{sim}(\ell,v^{*})\geq\tau$, where $\mathrm{Dice}_{3}$ is the Dice coefficient of character trigrams, $\mathrm{ED}$ is the edit distance, both computed on canonical forms, and $\tau=0.75$. Numeric literals, table and column names, and the clause structure are never modified, and every substitution is logged.

\subsubsection{Schema-grounded evidence}
\label{sec:sg}

The relational layer enters the prompt as foreign-key lines \sql{T1.C1 = T2.C2}. For the column-semantic layer, question phrases of up to four words are compared with table and column names after case folding, singular--plural folding, and five-character prefix matching. Following the linking relations of SGU-SQL \cite{zhang2025sgusql}, a phrase links to an element by exact-linking when the normalized forms coincide and by partial-linking when its words cover at least half of the words of the element name. Tied candidates are all kept, so that an ambiguous term remains visible. When the candidates of one phrase differ in whether the question contains a value stored in their column, each supported candidate is tagged, e.g., \sql{[question value 'bowel obstruction' exists in this column]}; the tag states the evidence but leaves the choice to the model. At most 12 links are rendered.

\subsection{Evidence-grounded CoT synthesis}
\label{sec:synthesis}

\begin{figure*}[t]
\centering
\includegraphics[width=\textwidth]{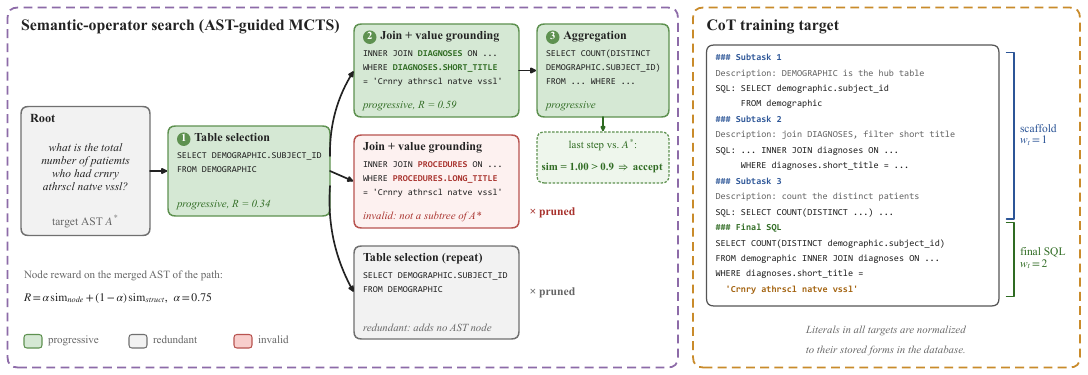}
\caption{Evidence-grounded CoT synthesis. Left: AST-guided MCTS expands progressive operators and prunes redundant and invalid ones. Right: the accepted path becomes a CoT whose tokens are weighted by 1, followed by the final SQL weighted by 2; literals take their stored forms in the database.}
\label{fig:sod}
\end{figure*}

For each training pair $(Q,Y^{*})$ whose reference executes, the teacher LLM DeepSeek-V4-Flash \cite{deepseek2026v4} proposes the next operator given the schema, the foreign-key hints of $\mathcal{G}$, $Y^{*}$, the preceding operators, and three demonstrations retrieved by AST similarity. Proposals are organized by MCTS \cite{browne2012mcts} with the upper confidence bound (UCB1) rule (50 simulations, depth at most 5). As in \cite{liao2026learnat}, a proposal is progressive if its AST is contained in the AST of $Y^{*}$ and extends the merged AST of its path, redundant if it adds nothing, and invalid otherwise. Only progressive nodes are expanded, with reward
\begin{equation}
R=\alpha\,\mathrm{sim}_{\mathrm{node}}(A_{\pi},A^{*})+(1-\alpha)\,\mathrm{sim}_{\mathrm{struct}}(A_{\pi},A^{*}),
\label{eq:reward}
\end{equation}
where $A_{\pi}$ is the merged AST of path $\pi$, $A^{*}$ is the AST of $Y^{*}$, $\mathrm{sim}_{\mathrm{node}}$ is a weighted Jaccard overlap of clause, operator, and operand nodes, $\mathrm{sim}_{\mathrm{struct}}=1-\mathrm{TED}/\max(|A_{\pi}|,|A^{*}|)$ with tree edit distance $\mathrm{TED}$, and $\alpha=0.75$. A path is accepted once its last step or its merged AST reaches a similarity above 0.9 to $A^{*}$ (Fig.~\ref{fig:sod}).

The search yields 7,129 verified trajectories, and 1,105 question-side variants that swap title-column cues or value surface forms while keeping the SQL unchanged give 8,234 samples. Every literal in the step fragments, the final SQL, and the descriptions is then normalized to its stored form in the value index, and GEM attaches the evidence block of each question.

\subsection{Grounded CoT distillation and inference}
\label{sec:distill}

The student learns to emit the CoT and then the final SQL under the token-weighted objective
\begin{equation}
\mathcal{L}=-\sum_{t}w_{t}\log p_{\theta}(y_{t}\mid y_{<t},x),
\label{eq:loss}
\end{equation}
where the prompt $x$ contains the schema, the foreign keys, the evidence block $E$, and the question, and $w_{t}$ is 1 for CoT tokens and 2 for final-SQL tokens. A two-phase curriculum trains on easy and medium samples first and then on all samples with hard samples oversampled, after which one further epoch trains on all samples without the curriculum. At inference, the student generates the CoT and the SQL in a single pass with GEM evidence in its prompt, without invoking the teacher, and value rewriting is applied to the final query.

%% ======================================================================
\section{Experiments}
\label{sec:exp}

We evaluated \method{} on the two datasets described in Section~\ref{sec:data}. All models are trained on the verified CoT samples and produce the CoT and the final SQL in a single decoding pass with GEM evidence in the prompt. %This section describes the evaluation setup and implementation details, and then reports quantitative results,qualitative results, and ablation studies.
On MIMICSQL, models are trained on the training split and evaluated on the 1,000 test questions of the official split \cite{wang2020treqs,pan2021medts,chen2026graph}. For cross-domain evaluation, the models trained on MIMICSQL are adapted to Spider with low-rank adaptation (LoRA) \cite{hu2022lora} on 832 training questions from 138 databases and evaluated on held-out Spider questions whose databases do not occur in the adaptation data; GEM is applied to Spider without modification.

%\subsection{Datasets}
%\label{sec:data}
%
%On MIMICSQL, models are trained on the training split and evaluated on the 1,000 test questions of the official split \cite{wang2020treqs,pan2021medts,chen2026graph}. For cross-domain evaluation, the models trained on MIMICSQL are adapted to Spider with low-rank adaptation (LoRA) \cite{hu2022lora} on 832 training questions from 138 databases and evaluated on held-out Spider questions whose databases do not occur in the adaptation data. And GEM is applied unchanged.

\subsection{Implementation details}
\label{sec:setup}

We adopted Qwen2.5-Coder-1.5B and Qwen2.5-Coder-7B \cite{hui2024qwen25coder} as backbones, fully fine-tuned with AdamW \cite{loshchilov2019adamw}, a peak learning rate of $2\times10^{-5}$ with cosine decay and 10\% warm-up, an effective batch size of 32, weight decay of 0.01, gradient clipping at 1.0, and bf16 precision for two curriculum epochs and one flat epoch. The Spider adapters (rank 16, scaling factor 32, dropout 0.05 on the attention projections) are trained for three epochs with a learning rate of $10^{-4}$. Decoding uses deterministic beam search with four beams and at most 640 new tokens.

\subsection{Evaluation metrics}
\label{sec:metric}

We adopted execution accuracy (EX), the fraction of predictions whose execution results match those of the reference queries \cite{yu2018spider} as our evaluation metric. Results are compared as sets of rows, invariant to column order, with string values compared case-insensitively. When a reference returns an empty result, a correct prediction cannot be distinguished from a trivially empty one. Following \cite{chen2026graph}, we therefore report EX under two conventions and their mean (average): \wempty{} evaluates all samples and counts an empty prediction as correct when the reference is empty, whereas \woempty{} excludes the samples with empty references, namely 5 in MIMICSQL and 54 in Spider.

The reliability of the CoT is measured on the raw generator output before value rewriting, following the consistency and factuality criteria of \cite{ye2022unreliability}. Let $A_{e}(\cdot)$ denote the elements of type $e\in\{$table, column, value, aggregation, join$\}$ in an SQL statement, where values are compared after case and whitespace normalization and joins as unordered pairs of key columns. With $S_{e}=\bigcup_{i}A_{e}(s_{i})$ the elements declared in the steps and $F_{e}=A_{e}(\hat{Y})$ those of the final query,
\begin{equation}
\mathrm{Cov}_{e}=\frac{|S_{e}\cap F_{e}|}{|S_{e}|},\qquad \mathrm{Trc}_{e}=\frac{|S_{e}\cap F_{e}|}{|F_{e}|},
\label{eq:consistency}
\end{equation}
averaged over the outputs with non-empty denominators. Coverage measures whether the decisions declared in the steps are carried into the final query, and traceability whether every element of the final query originates from a step. For factuality, value support is the fraction of equality conditions \sql{T.C = 'v'} in the steps whose literal is stored in \sql{T.C}, and schema support the fraction of column references in the steps that exist in the schema.

\subsection{Quantitative results}
\label{sec:quant}

\begin{table}[t]
\caption{The execution accuracy (\textbf{EX}) on MIMICSQL. General-purpose LLMs are queried with a few training demonstrations and without fine-tuning. Under \woempty{} and Average, \method{} obtains 0.943 (1.5B) and 0.956 (7B).}
\label{tab:mimic}
\footnotesize
\begin{tabular*}{\columnwidth}{@{\extracolsep{\fill}}lr@{}}
\toprule
Methods & Execution Accuracy\\
\midrule
Seq2Seq \cite{sutskever2014seq2seq} & 0.173\\
SQLNet \cite{xu2017sqlnet} & 0.260\\
PtrGen \cite{see2017ptrgen} & 0.292\\
Coarse2Fine \cite{dong2018coarse} & 0.496\\
TREQS \cite{wang2020treqs} & 0.654\\
MedTS \cite{pan2021medts} & 0.899\\
Graph-empowered (8B) \cite{chen2026graph} & 0.942\\
\midrule
\multicolumn{2}{@{}l}{\textit{General-purpose LLMs, few-shot prompting}}\\
GLM-4.7-Flash \cite{zai2026glm47flash} & 0.447\\
MiniMax-M3 \cite{minimax2026m3} & 0.754\\
\midrule
\method{} (1.5B) & \underline{0.943}\\
\method{} (7B) & \textbf{0.956}\\
\bottomrule
\end{tabular*}
\end{table}

\begin{table}[t]
\caption{The execution accuracy (\textbf{EX}) on Spider. Baseline values are those reported in \cite{chen2026graph}.}
\label{tab:spider}
\footnotesize
\begin{tabular*}{\columnwidth}{@{\extracolsep{\fill}}lr@{}}
\toprule
Methods & Execution Accuracy\\
\midrule
Global-GNN \cite{bogin2019global} & 0.474\\
RAT-SQL \cite{wang2020ratsql} & 0.572\\
Light RAT-SQL \cite{ndongala2023light} & 0.603\\
AuxNet + BART \cite{chen2026graph} & 0.626\\
BRIDGE$_{L}$ (ensemble) \cite{lin2020bridge} & 0.683\\
ChatGPT-SQL \cite{liu2023chatgpt} & 0.701\\
%DELLM$_{GPT\text{-}4}$ \cite{hong2024k2sql} & 0.717\\
%NatSQL + GAP \cite{gan2021natsql} & 0.733\\
%RASAT + PICARD \cite{qi2022rasat} & \textbf{0.755}\\
Graph-empowered (8B) \cite{chen2026graph} & 0.696\\
\midrule
\method{} (1.5B) & 0.510\\
\method{} (7B) & \textbf{0.712}\\
\bottomrule
\end{tabular*}
\end{table}

\begin{table*}[t]
\caption{Reliability of the CoT of \method, measured on the raw generator output (\%). Coverage and traceability follow Eq.~\eqref{eq:consistency}; traceability is identical for all five element types. Value and schema support are computed against the MIMICSQL database.}
\label{tab:consistency}
\footnotesize
\begin{tabular*}{\textwidth}{@{\extracolsep{\fill}}llcccccccc@{}}
\toprule
 & & & \multicolumn{5}{c}{Coverage} & & \\
\cmidrule(lr){4-8}
Dataset & Model & Parse rate & Tables & Columns & Values & Aggregations & Joins & Traceability & Value / schema support\\
\midrule
MIMICSQL & \method{} (1.5B) & 100.00 & 100.00 & 99.95 & 100.00 & 100.00 & 100.00 & 100.00 & 97.17 / 99.82\\
MIMICSQL & \method{} (7B) & 100.00 & 100.00 & 100.00 & 100.00 & 100.00 & 100.00 & 100.00 & 97.36 / 99.98\\
Spider & \method{} (1.5B) & 98.60 & 99.92 & 99.63 & 99.91 & 99.67 & 99.59 & 100.00 & --\\
Spider & \method{} (7B) & 99.91 & 99.89 & 99.96 & 100.00 & 99.86 & 100.00 & 100.00 & --\\
\bottomrule
\end{tabular*}
\end{table*}

We first compared \method{} with previous methods on MIMICSQL. The results are shown in Table~\ref{tab:mimic}. With a 1.5B student, \method{} reaches 0.943 EX, and the 7B student raises the accuracy to 0.956. Table~\ref{tab:spider} reports cross-domain results on Spider, where \method{} reaches 0.510 with 1.5B parameters and 0.712 with 7B parameters.

Table~\ref{tab:consistency} quantifies the reliability of the CoT. On MIMICSQL, every output follows the step format, traceability is 100\% for all element types, and coverage is at least 99.95\%, i.e., the final SQL contains nothing that the steps do not declare. The steps are also largely factual, with 97.2\% and 97.4\% of their value conditions naming stored values and over 99.8\% of their column references existing in the schema. On Spider, traceability remains 100\% and coverage at least 99.59\%; the 1.4\% of 1.5B outputs that miss the step format are truncated by the token budget.

\subsection{Qualitative results}
\label{sec:qual}

\begin{figure*}[t]
\centering
\footnotesize
\setlength{\fboxsep}{4pt}
\casepair{%
\textbf{(a) Table grounding} (prescription or procedure)\\[1pt]
\textit{Q:} find the number of patients who underwent drug detoxification.\\[1pt]
\evid{\textit{Value evidence:} \sql{PROCEDURES.SHORT\_TITLE = 'Drug detoxification'}; \sql{PROCEDURES.LONG\_TITLE = 'Drug detoxification'}}\\[2pt]
\textcolor{black!60}{/* CoT without GEM, step 2 */}\\
Join the \bad{PRESCRIPTIONS} table to DEMOGRAPHIC on HADM\_ID, identify the \bad{ROUTE} column, and filter for the value \bad{'DRUG detoxification'}.\\[2pt]
\textcolor{black!60}{/* \method, step 2 */}\\
Join the \good{PROCEDURES} table to DEMOGRAPHIC on HADM\_ID, identify the \good{SHORT\_TITLE} column, and filter for the value \good{'Drug detoxification'}.
}{%
\textbf{(b) Column grounding through a value tag}\\[1pt]
\textit{Q:} provide the number of patients who are diagnosed with bowel obstruction.\\[1pt]
\evid{\textit{Schema evidence:} \sql{'patients who are diagnosed' [Partial-Linking] -> DEMOGRAPHIC.DIAGNOSIS [question value 'bowel obstruction' exists in this column]}}\\[2pt]
\textcolor{black!60}{/* CoT without GEM, step 2 */}\\
Join the DIAGNOSES table to DEMOGRAPHIC on HADM\_ID, identify the \bad{SHORT\_TITLE} column, and filter for the value \bad{'Bowel obstruction'}.\\[2pt]
\textcolor{black!60}{/* \method, step 1 */}\\
Select the DEMOGRAPHIC table and identify the \good{DIAGNOSIS} column \ldots{} Filter patients whose diagnosis is \good{'BOWEL OBSTRUCTION'}.
}\\[5pt]
\casepair{%
\textbf{(c) Value grounding visible in the CoT}\\[1pt]
\textit{Q:} what is the number of patients who have transfer from hosp/extram as the admission location and are less than 59 years of age?\\[1pt]
\evid{\textit{Value evidence:} \sql{DEMOGRAPHIC.ADMISSION\_LOCATION = 'TRANSFER FROM HOSP/EXTRAM'}}\\[2pt]
\textcolor{black!60}{/* CoT without GEM, step 2 */}\\
Filter patients whose admission location is \bad{'transfer within this facility'} and whose age is less than 59 years.\\
\sql{... WHERE demographic.admission\_location = \bad{'TRSF WITHIN THIS FACILITY'} AND ...}\\[2pt]
\textcolor{black!60}{/* \method, step 2 */}\\
Filter patients whose admission location is \good{'TRANSFER FROM HOSP/EXTRAM'} and whose age is less than 59 years.
}{%
\textbf{(d) Incorrect query, consistent CoT} (\method)\\[1pt]
\textit{Q:} calulate the total number of patient who had icd9 code 5771\\[1pt]
\evid{\textit{Value evidence:} \sql{DIAGNOSES.ICD9\_CODE = '5771'}; \sql{PROCEDURES.ICD9\_CODE = '5771'}}\\[2pt]
\textcolor{black!60}{/* \method, step 2 */}\\
Join DEMOGRAPHIC to \bad{DIAGNOSES} on HADM\_ID and filter for ICD9\_CODE = '5771'\\[2pt]
\textcolor{black!60}{/* Reference query */}\\
\sql{... INNER JOIN \good{PROCEDURES} ON ... WHERE \good{PROCEDURES}.ICD9\_CODE = '5771'}\\[2pt]
\textcolor{black!60}{/* Stored titles of code 5771 */}\\
diagnosis: Chronic pancreatitis; procedure: Radical cystectomy
}
\caption{Visualization of qualitative results on MIMICSQL. Teal lines show the GEM evidence in the prompt; red marks a faulty decision and green the correct one. In (a)--(c), the step of the CoT model trained without GEM states its faulty decision in natural language, and \method{} returns the reference result. In (d), the query of \method{} is incorrect, but its step states which reading of the ambiguous code the query uses.}
\label{fig:qual}
\end{figure*}

On MIMICSQL, Seq2Seq \cite{sutskever2014seq2seq}, SQLNet \cite{xu2017sqlnet}, PtrGen \cite{see2017ptrgen}, and Coarse2Fine \cite{dong2018coarse} remain below 0.50 EX because they treat SQL as a token sequence or depend on fixed sketches. TREQS \cite{wang2020treqs} reaches 0.654 via controlled copying and value recovery, and MedTS \cite{pan2021medts} reaches 0.899 with a pre-trained encoder and grammar-based decoder. Graph-empowered Text-to-SQL \cite{chen2026graph} injects schema-graph embeddings into an 8B Llama~3.1 model \cite{grattafiori2024llama3} and attains 0.942; \method{} matches this with less than one fifth of the parameters while also returning a CoT. Few-shot prompting of general-purpose LLMs remains well below fine-tuning, as demonstrations do not convey database storage conventions. On Spider, the 7B student exceeds graph-empowered Text-to-SQL, and LoRA adaptation preserves the step format, suggesting that the distilled procedure transfers across domains. The 1.5B student is less accurate because 46.1\% of Spider questions contain no literal (Table~\ref{tab:stats}), so GEM has less evidence to offer.

Fig.~\ref{fig:qual} illustrates how the CoT supports inspection. In examples (a)--(c), the model trained without GEM restates its grounding error in the reasoning step, e.g., a different admission location in (c), so that the discrepancy is visible to a reader who does not know SQL; with GEM, \method{} decides correctly. Example (d) shows an incorrect query of \method{} itself. The code 5771 is stored both as a diagnosis and as a procedure. The reasoning step states that the query counts diagnoses, which a reviewer can check against the intended meaning. This aligns with observations that structured reasoning helps people examine model outputs and consider alternative interpretations \cite{alonso2026framing,shuai2025deepvis}. 

\subsection{Ablation study}
\label{sec:ablation}

\begin{table}[t]
\caption{Ablation on MIMICSQL with Qwen2.5-Coder-1.5B. Exec.: fraction of executable predictions; Trc.: traceability (\%); Val.: value support (\%). Settings without CoT have no reasoning steps to evaluate.}
\label{tab:ablation}
\footnotesize
\begin{tabular*}{\columnwidth}{@{\extracolsep{\fill}}lccccccc@{}}
\toprule
 & SFT & CoT & GEM & Exec. & EX & Trc. & Val.\\
\midrule
1 & & & & 0.562 & 0.231 & -- & --\\
2 & \cmark & & & 0.997 & 0.907 & -- & --\\
3 & \cmark & \cmark & & 0.977 & 0.870 & 100.00 & 88.86\\
4 & \cmark & \cmark & \cmark & 0.987 & \textbf{0.943} & 100.00 & 97.17\\
\bottomrule
\end{tabular*}
\end{table}

Table~\ref{tab:ablation} adds supervised fine-tuning (SFT), the CoT, and GEM in turn. All fine-tuned settings use the same 8,234 samples and training budget; the SQL-only setting keeps only the final SQL as target, and settings without GEM disable all its functionalities, including prompt evidence and value rewriting.

%\subsubsection{Untuned backbone}
%\label{sec:abl-raw}

Without fine-tuning, Qwen2.5-Coder-1.5B reaches 0.231 EX, and 43.8\% of its predictions fail to execute. Fine-tuning on the final SQL alone raises EX to 0.907 but returns no account of how the query is constructed. Adding the CoT lowers EX to 0.870. The paired outcomes confirm that this accuracy cost of explicit CoT is systematic: the CoT model answers 22 questions that the SQL-only model misses but fails on 59 that it answers; and McNemar's test \cite{mcnemar1947} rejects equal error rates ($p<10^{-4}$). Of its 130 incorrect queries, 122 select a wrong table (59), column (29), or literal value (34), i.e., the grounding decisions that the CoT exposes, and only 88.86\% of the value conditions in its steps name stored values. Adding GEM raises EX from 0.870 to 0.943 (+7.3 points). The value support rises from 88.86\% to 97.17\% while traceability stays at 100\%, so the gain comes with steps naming stored values rather than with any relaxation of the step format, and executable predictions recover from 0.977 to 0.987. GEM thus offsets the accuracy cost of explicit CoT, and the resulting model exceeds the SQL-only model by 3.6 points while returning an auditable reasoning trace. 

%\subsubsection{SFT without CoT}
%\label{sec:abl-sql}
%
%Fine-tuning on the final SQL alone raises EX to 0.907. This SQL-only model is accurate but returns no account of how its query was constructed.

%\subsubsection{SFT with CoT}
%\label{sec:abl-cot}
%
%Adding the CoT to the target lowers EX to 0.870. The paired outcomes confirm that this accuracy cost of explicit CoT is systematic: the CoT model answers 22 questions that the SQL-only model misses but fails on 59 that it answers, and McNemar's test \cite{mcnemar1947} rejects equal error rates ($p<10^{-4}$). Of its 130 incorrect queries, 122 select a wrong table (59), column (29), or literal value (34), i.e., exactly the grounding decisions that the CoT exposes, and only 88.86\% of the value conditions in its steps name stored values.

%\subsubsection{SFT with CoT and GEM}
%\label{sec:abl-gem}
%
%Adding GEM to the CoT model raises EX from 0.870 to 0.943, a gain of 7.3 points. The value support of the steps rises from 88.86\% to 97.17\% while traceability stays at 100\%, i.e., the gain comes with the steps naming stored values rather than with any relaxation of the step format, and the fraction of executable predictions recovers from 0.977 to 0.987. GEM thus removes the accuracy cost of explicit CoT observed, and the resulting model exceeds the SQL-only model by 3.6 points while additionally returning an auditable reasoning trace. 

%% ======================================================================
\section{Conclusion}
\label{sec:conclusion}

The study presents \method, a Text-to-SQL framework for EMRs whose structured CoT renders each grounding decision inspectable to medical professionals, and whose grounded evidence module uses value- and schema-level database evidence to offset the accuracy cost of explicit reasoning without trainable parameters, reference queries, or additional LLM calls. \method{} reaches execution accuracy of 0.943 and 0.956 on MIMICSQL with 1.5B and 7B generators, exceeding an 8B graph-injected LLM, while its reasoning steps remain consistent with the final SQL. Two limitations remain. On Spider, where fewer questions contain literals, GEM offers less evidence and the 1.5B model is markedly less accurate; the CoT is several times longer than the SQL alone, which increases decoding time. Future work will address temporal benchmarks such as EHRSQL \cite{lee2022ehrsql} and privacy-preserving deployment.

\section*{CRediT authorship contribution statement}
\textbf{Weikuan Qiu:} Conceptualization, Methodology, Software. \textbf{Chongzhao Xu:} Writing -- original draft, Visualization. \textbf{Xijian Zhuo:} Writing -- review \& editing. \textbf{Jinlian Ma:} Conceptualization, Supervision, Writing -- review \& editing.

\section*{Declaration of competing interests}
The authors declare that they have no known competing financial interests or personal relationships that could have appeared to influence the work reported in this paper.

\section*{Funding}
This research did not receive any specific grant from funding agencies in the public, commercial, or not-for-profit sectors.

\section*{Data availability}
MIMICSQL and Spider are publicly available. The MIMIC-III database underlying MIMICSQL requires credentialed access through PhysioNet. The code and the synthesized training data will be released upon publication.

\section*{Declaration of generative AI and AI-assisted technologies in the manuscript preparation process}
During the preparation of this work, the authors used Kimi AI to improve the language of the manuscript. After using this tool, the authors reviewed and edited the content as needed and take full responsibility for the content of the published article.

\section*{Acknowledgments}
The authors would like to express their gratitude to the authors of the MIMICSQL and Spider datasets for making their
data publicly available, and the developers of the LearNAT framework (\url{https://github.com/MrBlankness/LearNAT}) for providing well-documented, reproducible resources that supported the experiments on CoT training data synthesis in this work.

\bibliographystyle{elsarticle-num}
\bibliography{refs}

@article{johnson2016mimic,
  author  = {Johnson, Alistair E. W. and Pollard, Tom J. and Shen, Lu and Lehman, Li-wei H. and Feng, Mengling and Ghassemi, Mohammad and Moody, Benjamin and Szolovits, Peter and Celi, Leo Anthony and Mark, Roger G.},
  title   = {{MIMIC-III}, a freely accessible critical care database},
  journal = {Scientific Data},
  volume  = {3},
  pages   = {160035},
  year    = {2016}
}

@inproceedings{lee2022ehrsql,
  author    = {Lee, Gyubok and Hwang, Hyeonji and Bae, Seongsu and Kwon, Yeonsu and Shin, Woncheol and Yang, Seongjun and Seo, Minjoon and Kim, Jong-Yeup and Choi, Edward},
  title     = {{EHRSQL}: A practical text-to-{SQL} benchmark for electronic health records},
  booktitle = {Advances in Neural Information Processing Systems, Datasets and Benchmarks Track},
  volume    = {35},
  pages     = {15589--15601},
  year      = {2022}
}

@inproceedings{wang2020treqs,
  author    = {Wang, Ping and Shi, Tian and Reddy, Chandan K.},
  title     = {Text-to-{SQL} generation for question answering on electronic medical records},
  booktitle = {Proceedings of The Web Conference 2020},
  pages     = {350--361},
  year      = {2020}
}

@article{pan2021medts,
  author  = {Pan, Youcheng and Wang, Chenghao and Hu, Baotian and Xiang, Yang and Wang, Xiaolong and Chen, Qingcai and Chen, Junjie and Du, Jingcheng},
  title   = {A {BERT}-based generation model to transform medical texts to {SQL} queries for electronic medical records: Model development and validation},
  journal = {JMIR Medical Informatics},
  volume  = {9},
  number  = {12},
  pages   = {e32698},
  year    = {2021}
}

@article{kim2020nl2sql,
  author  = {Kim, Hyeonji and So, Byeong-Hoon and Han, Wook-Shin and Lee, Hongrae},
  title   = {Natural language to {SQL}: Where are we today?},
  journal = {Proceedings of the VLDB Endowment},
  volume  = {13},
  number  = {10},
  pages   = {1737--1750},
  year    = {2020}
}

@inproceedings{yu2018spider,
  author    = {Yu, Tao and Zhang, Rui and Yang, Kai and Yasunaga, Michihiro and Wang, Dongxu and Li, Zifan and Ma, James and Li, Irene and Yao, Qingning and Roman, Shanelle and Zhang, Zilin and Radev, Dragomir},
  title     = {Spider: A large-scale human-labeled dataset for complex and cross-domain semantic parsing and text-to-{SQL} task},
  booktitle = {Proceedings of the 2018 Conference on Empirical Methods in Natural Language Processing},
  pages     = {3911--3921},
  year      = {2018}
}

@inproceedings{li2023bird,
  author    = {Li, Jinyang and Hui, Binyuan and Qu, Ge and Yang, Jiaxi and Li, Binhua and Li, Bowen and Wang, Bailin and Qin, Bowen and Cao, Rongyu and Geng, Ruiying and others},
  title     = {Can {LLM} already serve as a database interface? {A} big bench for large-scale database grounded text-to-{SQLs}},
  booktitle = {Advances in Neural Information Processing Systems},
  volume    = {36},
  pages     = {42330--42357},
  year      = {2023}
}

@article{liu2025survey,
  author  = {Liu, Xinyu and Shen, Shuyu and Li, Boyan and Ma, Peixian and Jiang, Runzhi and Zhang, Yuxin and Fan, Ju and Li, Guoliang and Tang, Nan and Luo, Yuyu},
  title   = {A survey of text-to-{SQL} in the era of {LLMs}: Where are we, and where are we going?},
  journal = {IEEE Transactions on Knowledge and Data Engineering},
  volume  = {37},
  number  = {10},
  pages   = {5735--5754},
  year    = {2025}
}

@inproceedings{wei2022flan,
  author    = {Wei, Jason and Bosma, Maarten and Zhao, Vincent Y. and Guu, Kelvin and Yu, Adams Wei and Lester, Brian and Du, Nan and Dai, Andrew M. and Le, Quoc V.},
  title     = {Finetuned language models are zero-shot learners},
  booktitle = {International Conference on Learning Representations},
  year      = {2022}
}

@article{hong2025survey,
  author  = {Hong, Zijin and Yuan, Zheng and Zhang, Qinggang and Chen, Hao and Dong, Junnan and Huang, Feiran and Huang, Xiao},
  title   = {Next-generation database interfaces: A survey of {LLM}-based text-to-{SQL}},
  journal = {IEEE Transactions on Knowledge and Data Engineering},
  volume    = {37},
  number    = {12},
  pages     = {7328--7345},
  year    = {2025}
}

@article{chen2026graph,
  author  = {Chen, Qian and Peng, Jun and Song, Baiyang and Zhou, Yiyi and Ji, Rongrong},
  title   = {Graph-empowered Text-to-{SQL} generation on Electronic Medical Records},
  journal = {Pattern Recognition},
  volume  = {169},
  pages   = {111800},
  year    = {2026}
}

@article{wei2022cot,
  author  = {Wei, Jason and Wang, Xuezhi and Schuurmans, Dale and Bosma, Maarten and Ichter, Brian and Xia, Fei and Chi, Ed and Le, Quoc V. and Zhou, Denny},
  title   = {Chain-of-thought prompting elicits reasoning in large language models},
  journal = {Advances in Neural Information Processing Systems},
  volume  = {35},
  pages   = {24824--24837},
  year    = {2022}
}

@inproceedings{pampari2018emrqa,
  author    = {Pampari, Anusri and Raghavan, Preethi and Liang, Jennifer and Peng, Jian},
  title     = {{emrQA}: A large corpus for question answering on electronic medical records},
  booktitle = {Proceedings of the 2018 Conference on Empirical Methods in Natural Language Processing},
  pages     = {2357--2368},
  year      = {2018}
}

@inproceedings{devlin2019bert,
  author    = {Devlin, Jacob and Chang, Ming-Wei and Lee, Kenton and Toutanova, Kristina},
  title     = {{BERT}: Pre-training of deep bidirectional transformers for language understanding},
  booktitle = {Proceedings of the 2019 Conference of the North American Chapter of the Association for Computational Linguistics},
  pages     = {4171--4186},
  year      = {2019}
}

@article{tarbell2023medical,
  author  = {Tarbell, Richard and Choo, Kim-Kwang Raymond and Dietrich, Glenn and Rios, Anthony},
  title   = {Towards Understanding the Generalization of Medical {Text-to-SQL} Models and Datasets},
  journal = {AMIA Annual Symposium Proceedings},
  volume  = {2023},
  pages   = {669--678},
  year    = {2023}
}

@article{androutsopoulos1995nlidb,
  author  = {Androutsopoulos, Ion and Ritchie, Graeme D. and Thanisch, Peter},
  title   = {Natural language interfaces to databases -- An introduction},
  journal = {Natural Language Engineering},
  volume  = {1},
  number  = {1},
  pages   = {29--81},
  year    = {1995}
}

@article{saias2012binli,
  author  = {Saias, Jos{\'e} and Quaresma, Paulo and Salgueiro, Pedro and Santos, Tiago},
  title   = {{BINLI}: An ontology-based natural language interface for multidimensional data analysis},
  journal = {Intelligent Information Management},
  volume  = {4},
  number  = {5},
  pages   = {225--230},
  year    = {2012}
}

@article{xu2017sqlnet,
  author  = {Xu, Xiaojun and Liu, Chang and Song, Dawn},
  title   = {{SQLNet}: Generating structured queries from natural language without reinforcement learning},
  journal = {arXiv preprint arXiv:1711.04436},
  year    = {2017}
}

@article{choi2021ryansql,
  author  = {Choi, DongHyun and Shin, Myeong Cheol and Kim, EungGyun and Shin, Dong Ryeol},
  title   = {{RYANSQL}: Recursively applying sketch-based slot fillings for complex text-to-{SQL} in cross-domain databases},
  journal = {Computational Linguistics},
  volume  = {47},
  number  = {2},
  pages   = {309--332},
  year    = {2021}
}

@inproceedings{dong2018coarse,
  author    = {Dong, Li and Lapata, Mirella},
  title     = {Coarse-to-fine decoding for neural semantic parsing},
  booktitle = {Proceedings of the 56th Annual Meeting of the Association for Computational Linguistics},
  pages     = {731--742},
  year      = {2018}
}

@inproceedings{bogin2019global,
  author    = {Bogin, Ben and Gardner, Matt and Berant, Jonathan},
  title     = {Global reasoning over database structures for text-to-{SQL} parsing},
  booktitle = {Proceedings of the 2019 Conference on Empirical Methods in Natural Language Processing},
  pages     = {3657--3662},
  year      = {2019}
}

@inproceedings{bogin2019gnn,
  author    = {Bogin, Ben and Berant, Jonathan and Gardner, Matt},
  title     = {Representing schema structure with graph neural networks for text-to-{SQL} parsing},
  booktitle = {Proceedings of the 57th Annual Meeting of the Association for Computational Linguistics},
  pages     = {4560--4565},
  year      = {2019}
}

@inproceedings{wang2020ratsql,
  author    = {Wang, Bailin and Shin, Richard and Liu, Xiaodong and Polozov, Oleksandr and Richardson, Matthew},
  title     = {{RAT-SQL}: Relation-aware schema encoding and linking for text-to-{SQL} parsers},
  booktitle = {Proceedings of the 58th Annual Meeting of the Association for Computational Linguistics},
  pages     = {7567--7578},
  year      = {2020}
}

@inproceedings{li2023graphix,
  author    = {Li, Jinyang and Hui, Binyuan and Cheng, Reynold and Qin, Bowen and Ma, Chenhao and Huo, Nan and Huang, Fei and Du, Wenyu and Si, Luo and Li, Yongbin},
  title     = {{Graphix-T5}: Mixing pre-trained transformers with graph-aware layers for text-to-{SQL} parsing},
  booktitle = {Proceedings of the AAAI Conference on Artificial Intelligence},
  volume    = {37},
  pages     = {13076--13084},
  year      = {2023}
}

@article{liu2019roberta,
  author  = {Liu, Yinhan and Ott, Myle and Goyal, Naman and Du, Jingfei and Joshi, Mandar and Chen, Danqi and Levy, Omer and Lewis, Mike and Zettlemoyer, Luke and Stoyanov, Veselin},
  title   = {{RoBERTa}: A robustly optimized {BERT} pretraining approach},
  journal = {arXiv preprint arXiv:1907.11692},
  year    = {2019}
}

@inproceedings{li2023resdsql,
  author    = {Li, Haoyang and Zhang, Jing and Li, Cuiping and Chen, Hong},
  title     = {{RESDSQL}: Decoupling schema linking and skeleton parsing for text-to-{SQL}},
  booktitle = {Proceedings of the AAAI Conference on Artificial Intelligence},
  volume    = {37},
  pages     = {13067--13075},
  year      = {2023}
}

@article{gao2024dailsql,
  author  = {Gao, Dawei and Wang, Haibin and Li, Yaliang and Sun, Xiuyu and Qian, Yichen and Ding, Bolin and Zhou, Jingren},
  title   = {Text-to-{SQL} empowered by large language models: A benchmark evaluation},
  journal = {Proceedings of the VLDB Endowment},
  volume  = {17},
  number  = {5},
  pages   = {1132--1145},
  year    = {2024}
}

@inproceedings{lee2025safe,
  author    = {Lee, Jimin and Baek, Ingeol and Kim, Byeongjeong and Bae, Hyunkyung and Lee, Hwanhee},
  title     = {{SAFE-SQL}: Self-augmented in-context learning with fine-grained example selection for text-to-{SQL}},
  booktitle = {Proceedings of the 2025 Conference on Empirical Methods in Natural Language Processing},
  pages     = {19023--19035},
  year      = {2025}
}

@article{pourreza2023dinsql,
  author  = {Pourreza, Mohammadreza and Rafiei, Davood},
  title   = {{DIN-SQL}: Decomposed in-context learning of text-to-{SQL} with self-correction},
  journal = {Advances in Neural Information Processing Systems},
  volume  = {36},
  pages     = {36339--36348},
  year    = {2023}
}

@inproceedings{zhou2023least,
  author    = {Zhou, Denny and Sch{\"a}rli, Nathanael and Hou, Le and Wei, Jason and Scales, Nathan and Wang, Xuezhi and Schuurmans, Dale and Cui, Claire and Bousquet, Olivier and Le, Quoc and Chi, Ed},
  title     = {Least-to-most prompting enables complex reasoning in large language models},
  booktitle = {International Conference on Learning Representations},
  year      = {2023}
}

@inproceedings{qu2024tasql,
  author    = {Qu, Ge and Li, Jinyang and Li, Bowen and Qin, Bowen and Huo, Nan and Ma, Chenhao and Cheng, Reynold},
  title     = {Before generation, align it! {A} novel and effective strategy for mitigating hallucinations in text-to-{SQL} generation},
  booktitle = {Findings of the Association for Computational Linguistics: ACL 2024},
  pages     = {5456--5471},
  year      = {2024}
}

@article{talaei2024chess,
  author  = {Talaei, Shayan and Pourreza, Mohammadreza and Chang, Yu-Chen and Mirhoseini, Azalia and Saberi, Amin},
  title   = {{CHESS}: Contextual harnessing for efficient {SQL} synthesis},
  journal = {arXiv preprint arXiv:2405.16755},
  year    = {2024}
}

@inproceedings{pourreza2025chase,
  author    = {Pourreza, Mohammadreza and Li, Hailong and Sun, Ruoxi and Chung, Yeounoh and Talaei, Shayan and Kakkar, Gaurav Tarlok and Gan, Yu and Saberi, Amin and Ozcan, Fatma and Arik, Sercan O.},
  title     = {{CHASE-SQL}: Multi-path reasoning and preference optimized candidate selection in text-to-{SQL}},
  booktitle = {International Conference on Learning Representations},
  volume    = {2025},
  pages     = {60385--60415},
  year      = {2025}
}

@article{li2025omnisql,
  author  = {Li, Haoyang and Wu, Shang and Zhang, Xiaokang and Huang, Xinmei and Zhang, Jing and Jiang, Fuxin and Wang, Shuai and Zhang, Tieying and Chen, Jianjun and Shi, Rui and others},
  title   = {{OmniSQL}: Synthesizing high-quality text-to-{SQL} data at scale},
  journal = {Proceedings of the VLDB Endowment},
  volume  = {18},
  number  = {11},
  pages   = {4695--4709},
  year    = {2025}
}

@article{sheng2025basesql,
  author  = {Sheng, Lei and Xu, Shuai-Shuai and Xie, Wei},
  title   = {{BASE-SQL}: A powerful open source text-to-{SQL} baseline approach},
  journal = {arXiv preprint arXiv:2502.10739},
  year    = {2025}
}

@inproceedings{zhang2025spft,
  author    = {Zhang, Yuhao and Duan, Shaoming and Su, Jinhang and Liu, Chuanyi and Han, Peiyi},
  title     = {{SPFT-SQL}: Enhancing large language model for text-to-{SQL} parsing by self-play fine-tuning},
  booktitle = {Findings of the Association for Computational Linguistics: EMNLP 2025},
  pages     = {1092--1110},
  year      = {2025}
}

@inproceedings{liao2026learnat,
  author    = {Liao, Weibin and Gao, Xin and Jia, Tianyu and Qiu, Rihong and Zhu, Yifan and Lin, Yang and Ma, Xinyu and Zhao, Junfeng and Wang, Yasha},
  title     = {{LearNAT}: Learning {NL2SQL} with {AST}-guided task decomposition for large language models},
  booktitle = {International Conference on Learning Representations},
  volume    = {2026},
  pages     = {87420--87452},
  year      = {2026}
}

@article{browne2012mcts,
  author  = {Browne, Cameron B. and Powley, Edward and Whitehouse, Daniel and Lucas, Simon M. and Cowling, Peter I. and Rohlfshagen, Philipp and Tavener, Stephen and Perez, Diego and Samothrakis, Spyridon and Colton, Simon},
  title   = {A survey of {Monte Carlo} tree search methods},
  journal = {IEEE Transactions on Computational Intelligence and AI in Games},
  volume  = {4},
  number  = {1},
  pages   = {1--43},
  year    = {2012}
}

@article{rafailov2023dpo,
  author  = {Rafailov, Rafael and Sharma, Archit and Mitchell, Eric and Ermon, Stefano and Manning, Christopher D. and Finn, Chelsea},
  title   = {Direct preference optimization: Your language model is secretly a reward model},
  journal = {Advances in Neural Information Processing Systems},
  volume  = {36},
  pages     = {53728--53741},
  year    = {2023}
}

@inproceedings{lin2020bridge,
  author    = {Lin, Xi Victoria and Socher, Richard and Xiong, Caiming},
  title     = {Bridging textual and tabular data for cross-domain text-to-{SQL} semantic parsing},
  booktitle = {Findings of the Association for Computational Linguistics: EMNLP 2020},
  pages     = {4870--4888},
  year      = {2020}
}

@article{deepseek2026v4,
  author  = {{DeepSeek-AI}},
  title   = {{DeepSeek-V4}: Towards highly efficient million-token context intelligence},
  journal = {arXiv preprint arXiv:2606.19348},
  year    = {2026}
}

@article{hui2024qwen25coder,
  author  = {Hui, Binyuan and Yang, Jian and Cui, Zeyu and Yang, Jiaxi and Liu, Dayiheng and Zhang, Lei and Liu, Tianyu and Zhang, Jiajun and Yu, Bowen and Lu, Keming and others},
  title   = {{Qwen2.5-Coder} technical report},
  journal = {arXiv preprint arXiv:2409.12186},
  year    = {2024}
}

@inproceedings{loshchilov2019adamw,
  author    = {Loshchilov, Ilya and Hutter, Frank},
  title     = {Decoupled weight decay regularization},
  booktitle = {International Conference on Learning Representations},
  year      = {2019}
}

@article{sutskever2014seq2seq,
  author  = {Sutskever, Ilya and Vinyals, Oriol and Le, Quoc V.},
  title   = {Sequence to sequence learning with neural networks},
  journal = {Advances in Neural Information Processing Systems},
  volume  = {27},
  pages     = {3104--3112},
  year    = {2014}
}

@inproceedings{see2017ptrgen,
  author    = {See, Abigail and Liu, Peter J. and Manning, Christopher D.},
  title     = {Get to the point: Summarization with pointer-generator networks},
  booktitle = {Proceedings of the 55th Annual Meeting of the Association for Computational Linguistics},
  pages     = {1073--1083},
  year      = {2017}
}

@misc{zai2026glm47flash,
  author       = {{Z.ai}},
  title        = {{GLM-4.7-Flash}},
  howpublished = {Model card, \url{https://huggingface.co/zai-org/GLM-4.7-Flash}},
  year         = {2026}
}

@misc{minimax2026m3,
  author       = {{MiniMax}},
  title        = {{MiniMax-M3}},
  howpublished = {Model card, \url{https://huggingface.co/MiniMaxAI/MiniMax-M3}; see also arXiv:2606.13392},
  year         = {2026}
}

@article{ndongala2023light,
  author  = {Ndongala, Nathan Manzambi},
  title   = {Light {RAT-SQL}: A {RAT-SQL} with more abstraction and less embedding of pre-existing relations},
  journal = {Texila International Journal of Academic Research},
  volume  = {10},
  number  = {2},
  pages   = {1--11},
  year    = {2023}
}

@article{liu2023chatgpt,
  author  = {Liu, Aiwei and Hu, Xuming and Wen, Lijie and Yu, Philip S.},
  title   = {A comprehensive evaluation of {ChatGPT}'s zero-shot text-to-{SQL} capability},
  journal = {arXiv preprint arXiv:2303.13547},
  year    = {2023}
}

@article{grattafiori2024llama3,
  author  = {Grattafiori, Aaron and Dubey, Abhimanyu and Jauhri, Abhinav and others},
  title   = {The {Llama 3} herd of models},
  journal = {arXiv preprint arXiv:2407.21783},
  year    = {2024}
}

@article{mcnemar1947,
  author  = {McNemar, Quinn},
  title   = {Note on the sampling error of the difference between correlated proportions or percentages},
  journal = {Psychometrika},
  volume  = {12},
  number  = {2},
  pages   = {153--157},
  year    = {1947}
}

@inproceedings{jacovi2020faithful,
  author    = {Jacovi, Alon and Goldberg, Yoav},
  title     = {Towards faithfully interpretable {NLP} systems: How should we define and evaluate faithfulness?},
  booktitle = {Proceedings of the 58th Annual Meeting of the Association for Computational Linguistics},
  pages     = {4198--4205},
  year      = {2020}
}

@article{turpin2023unfaithful,
  author  = {Turpin, Miles and Michael, Julian and Perez, Ethan and Bowman, Samuel R.},
  title   = {Language models don't always say what they think: Unfaithful explanations in chain-of-thought prompting},
  journal = {Advances in Neural Information Processing Systems},
  volume  = {36},
  pages     = {74952--74965},
  year    = {2023}
}

@article{ye2022unreliability,
  author  = {Ye, Xi and Durrett, Greg},
  title   = {The unreliability of explanations in few-shot prompting for textual reasoning},
  journal = {Advances in Neural Information Processing Systems},
  volume  = {35},
  pages     = {30378--30392},
  year    = {2022}
}

@inproceedings{lyu2023faithful,
  author    = {Lyu, Qing and Havaldar, Shreya and Stein, Adam and Zhang, Li and Rao, Delip and Wong, Eric and Apidianaki, Marianna and Callison-Burch, Chris},
  title     = {Faithful chain-of-thought reasoning},
  booktitle = {Proceedings of the 13th International Joint Conference on Natural Language Processing and the 3rd Conference of the Asia-Pacific Chapter of the Association for Computational Linguistics},
  pages     = {305--329},
  year      = {2023}
}

@inproceedings{elgohary2020speak,
  author    = {Elgohary, Ahmed and Hosseini, Saghar and Hassan Awadallah, Ahmed},
  title     = {Speak to your parser: Interactive text-to-{SQL} with natural language feedback},
  booktitle = {Proceedings of the 58th Annual Meeting of the Association for Computational Linguistics},
  pages     = {2065--2077},
  year      = {2020}
}

@inproceedings{tian2023steps,
  author    = {Tian, Yuan and Zhang, Zheng and Ning, Zheng and Li, Toby Jia-Jun and Kummerfeld, Jonathan K. and Zhang, Tianyi},
  title     = {Interactive text-to-{SQL} generation via editable step-by-step explanations},
  booktitle = {Proceedings of the 2023 Conference on Empirical Methods in Natural Language Processing},
  pages     = {16149--16166},
  year      = {2023}
}

@inproceedings{hsieh2023distilling,
  author    = {Hsieh, Cheng-Yu and Li, Chun-Liang and Yeh, Chih-kuan and Nakhost, Hootan and Fujii, Yasuhisa and Ratner, Alex and Krishna, Ranjay and Lee, Chen-Yu and Pfister, Tomas},
  title     = {Distilling step-by-step! {O}utperforming larger language models with less training data and smaller model sizes},
  booktitle = {Findings of the Association for Computational Linguistics: ACL 2023},
  pages     = {8003--8017},
  year      = {2023}
}

@article{li2024codes,
  author  = {Li, Haoyang and Zhang, Jing and Liu, Hanbing and Fan, Ju and Zhang, Xiaokang and Zhu, Jun and Wei, Renjie and Pan, Hongyan and Li, Cuiping and Chen, Hong},
  title   = {{CodeS}: Towards building open-source language models for text-to-{SQL}},
  journal = {Proceedings of the ACM on Management of Data},
  volume  = {2},
  number  = {3},
  pages   = {127:1--127:28},
  year    = {2024}
}

@inproceedings{hu2022lora,
  author    = {Hu, Edward J. and Shen, Yelong and Wallis, Phillip and Allen-Zhu, Zeyuan and Li, Yuanzhi and Wang, Shean and Wang, Lu and Chen, Weizhu},
  title     = {{LoRA}: Low-rank adaptation of large language models},
  booktitle = {International Conference on Learning Representations},
  year      = {2022}
}

@article{shuai2025deepvis,
  author  = {Shuai, Zhihao and Li, Boyan and Yan, Siyu and Luo, Yuyu and Yang, Weikai},
  title   = {{DeepVIS}: Bridging natural language and data visualization through step-wise reasoning},
  journal = {IEEE Transactions on Visualization and Computer Graphics},
  volume    = {32},
  number    = {1},
  pages     = {868--878},
  year      = {2026},
  doi     = {10.1109/TVCG.2025.3634645}
}

@inproceedings{zhang2025sgusql,
  author    = {Zhang, Qinggang and Chen, Hao and Dong, Junnan and Chen, Shengyuan and Huang, Feiran and Huang, Xiao},
  title     = {Structure-guided large language models for text-to-{SQL} generation},
  booktitle = {Proceedings of the 42nd International Conference on Machine Learning},
  series    = {Proceedings of Machine Learning Research},
  volume    = {267},
  pages     = {74671--74691},
  year      = {2025}
}

@inproceedings{sheng2025slmsql,
  author    = {Sheng, Lei and Xu, Shuai-Shuai},
  title     = {{SLM-SQL}: An exploration of small language models for text-to-{SQL}},
  booktitle = {Findings of the Association for Computational Linguistics: IJCNLP-AACL 2025},
  pages     = {1497--1512},
  year      = {2025}
}

@inproceedings{hoang2026finersql,
  author    = {Hoang, Thanh Dat and Huynh, Thanh Trung and Weidlich, Matthias and Nguyen, Thanh Tam and Chen, Tong and Yin, Hongzhi and Nguyen, Quoc Viet Hung},
  title     = {Boosting small language models for text-to-{SQL} with fine-grained execution feedback and cost-efficient rewards},
  booktitle = {Proceedings of the IEEE 42nd International Conference on Data Engineering (ICDE)},
  pages     = {1--14},
  year      = {2026}
}

@inproceedings{alonso2026framing,
  author    = {Alonso del Barrio, David and Wen, Jing and Gatica-Perez, Daniel},
  title     = {Framing migration news with {LLMs}: Structured {CoT} as a support for human interpretation},
  booktitle = {Proceedings of the ACM SIGCAS/SIGCHI Conference on Computing and Sustainable Societies (COMPASS)},
  pages     = {199--210},
  year      = {2026},
  doi       = {10.1145/3811242.3819110}
}

@inproceedings{zhou2026integrity,
  author    = {Zhou, Jiling and Adeseye, Aisvarya and Virtanen, Seppo and Hakkala, Antti and Isoaho, Jouni},
  title     = {Strengthening human-centric chain-of-thought reasoning integrity in {LLMs} via a structured prompt framework},
  booktitle = {Intelligent Systems and Applications: Proceedings of the 2026 Intelligent Systems Conference (IntelliSys)},
  series    = {Lecture Notes in Networks and Systems},
  publisher = {Springer},
  pages     = {174--195},
  year      = {2026}
}

\end{document}